\documentclass{article}
\pdfoutput=1

\usepackage[nonatbib,preprint]{neurips_2019}

\usepackage[numbers]{natbib}

\usepackage{mathptmx}      

\usepackage[utf8]{inputenc} 
\usepackage[T1]{fontenc}    
\usepackage{hyperref}       
\usepackage{url}            
\usepackage{booktabs}       
\usepackage{amsfonts}       
\usepackage{nicefrac}       
\usepackage{microtype}      
\usepackage{amsmath}
\usepackage{amssymb}
\usepackage{amsthm}
\usepackage{subfigure}
\usepackage{graphicx}
\usepackage{bm}

\theoremstyle{plain}

\newtheorem{proposition}{Proposition}

\newtheorem{corollary}{Corollary}
\newtheorem{example}{Example}

\theoremstyle{definition}
\newtheorem{remark}{Remark}

\newtheorem{definition}{Definition}

\def\citet{\citep}

\newcommand{\eps}{\varepsilon}

\def\m#1{\ensuremath{\mathtt{#1}}}
\def\v#1{\ensuremath{\mathbf{#1}}}
\def\mI{{\m I}}

\def\mA{{\m A}}
\def\mB{{\m B}}

\def\tr{^\top}

\providecommand{\bbeta}[0]{\ensuremath{\bm{\beta}}}

\providecommand{\norm}[1]{\lVert#1\rVert}
\providecommand{\Norm}[1]{\ensuremath{\left\lVert#1\right\rVert}}

\providecommand{\homogvec}[1]{\binom{n}{1}}

\providecommand{\cL}{\ensuremath{\mathcal{L}}}
\providecommand{\cE}{\ensuremath{\mathcal{E}}}

\begin{document}

\title{Bilevel Programs Meet Deep Learning: \\ A Unifying View on Inference Learning Methods}

\author{%
  Christopher Zach  \\
  Chalmers University of Technology\\
  Gothenburg, Sweden \\
  \texttt{christopher.m.zach@gmail.com}
}

\maketitle

\begin{abstract}
  In this work we unify a number of \emph{inference learning} methods, that
  are proposed in the literature as alternative training algorithms to the
  ones based on regular error back-propagation. These inference learning
  methods were developed with very diverse motivations, mainly aiming to
  enhance the biological plausibility of deep neural networks and to improve
  the intrinsic parallelism of training methods. We show that these
  superficially very different methods can all be obtained by successively
  applying a particular reformulation of bilevel optimization programs. As a
  by-product it becomes also evident that all considered inference learning
  methods include back-propagation as a special case, and therefore at least
  approximate error back-propagation in typical settings.  Finally, we propose
  \emph{Fenchel back-propagation}, that replaces the propagation of
  infinitesimal corrections performed in standard back-propagation with finite
  targets as the learning signal. Fenchel back-propagation can therefore be
  seen as an instance of learning via explicit \emph{target propagation}.
\end{abstract}

\section{Introduction}
\label{sec:intro}

In this work we analyze several instances of \emph{inference learning} methods
and how these approaches can be obtained as suitable reformulations of deeply
nested minimization problems. We use inference learning~\citep{song2020can} as
an umbrella term for all learning methods based on explicit optimization
(inference) of hidden unit activations. Inference learning includes methods
such as layered contrastive Hebbian learning (CHL,~\citet{xie2003equivalence}),
equilibrium propagation~\citep{scellier2017equilibrium}, lifted neural
networks~\citep{carreira2014distributed,gu2018fenchel,li2018lifted}, and
predictive coding networks
(PCN,~\citet{whittington2017approximation,song2020can}). In a nutshell, all
these methods can be interpreted as certain mathematical reformulations of
deeply nested optimization problems. Specifically, learning in layered
feed-forward neural network can be stated as the following nested optimization
problem,
\begin{align}
  \min_{\theta_0,\dotsc,\theta_{L-1}} \ell(z_L^*) \;\; \text{s.t. } \qquad z_1^* = \arg\min_{z_1} E_1,(x, z_1; \theta_0)
  \qquad z_{k+1}^* = \arg\min_{z_{k+1}} E_k(z_k^*, z_{k+1};\theta_k)
  \label{eq:DNN}
\end{align}
for $k=1,\dotsc,L$.
The values $(z_k^*)_{k=1}^L$ are usually refered to as the network activations
determined in a forward pass, and the collection
$(\theta_0,\dotsc,\theta_{L-1})$ are the trainable network parameters.
The conversion between a DNN architecture and Eq.~\ref{eq:DNN} is not unique.
E.g., for ReLU-based DNNs the choice for $E_k$ can be
$E_k(z_k,z_{k+1};\theta_k) =
\Norm{z_{k+1}-[\theta_kz_k]_+}^2$~\citep{carreira2014distributed} or
$E_k(z_k,z_{k+1};\theta_k) = \Norm{z_{k+1}-\theta_kz_k}^2 + \imath_{\ge
  0}(z_{k+1})$~\citep{zhang2017convergent}, where
$[\cdot]_+:=\max\{0,\cdot\}$ and $\imath_C$ is the extended valued
characteristic function of a set $C$. In our work it will become clear, that
the differences between inference learning methods is largely induced by
different (but equivalent) choices for the underlying nested optimization
problems and by selected direction for processing the nesting levels (top-down
or bottom-up). Since all these approaches yield the standard back-propagation
(BP) method at least as a limit case, understanding the connections between
the inference learning methods may also provide insights into regular
back-propagation.
In particular, some inference learning methods can be interpreted as instances
of target
propagation~\citep{bengio2014auto,lee2015difference,bengio2015towards,ororbia2019biologically,meulemans2020theoretical}---where
desired neural activations instead of (infinitesimal) error signals are
propagated through the network---and as the basis for learning methods using
layer-local losses (including~\cite{frerix2017proximal,amid2021locoprop}),
hence this work may also shed new light on those methods.

\begin{table*}[t]
  \centering
  \begin{tabular}{ccccc}
    Method & $E_k$ & Direction & Linearized outer loss & Inference \\ \hline
    CHL & Fenchel & Bottom-up & No & Layer-convex\\
    Convex CHL & Proximal & Bottom-up & No & Convex \\
    FLN \& LPOM & Fenchel/Proximal & Top-down & No & Layer-Convex \\
    MAC & Penalizer & Any & No & Non-convex \\
    BP \& PCN & Penalizer & Top-down & Yes & Closed-form \\
    Fenchel BP & Fenchel/Proximal & Top-down & Yes & Closed-form
  \end{tabular}
  \caption{A taxonomy of inference learning methods unified in this work:
    Contrastive Hebbian learning (CHL,~\citet{xie2003equivalence}) and convex
    CHL~\citep{zach2019contrastive}, Fenchel lifted networks
    (FLN,~\citet{gu2018fenchel}) and lifted proximal operator machines
    (LPOM,~\citet{li2018lifted}), method of auxiliary coordinates
    (MAC,~\citet{carreira2014distributed}), back-propagation (BP) and
    predictive coding networks
    (PCN,~\citet{whittington2017approximation,song2020can}). Fenchel
    back-propagation (Fenchel BP) is a generalization of standard
    back-propagation discussed in Section~\ref{sec:FenchelBP}.}
  \label{tab:taxonomy}
\end{table*}

Table~\ref{tab:taxonomy} summarizes the relations between inference learning
methods. As it can be seen, the methods differ in the choice of $E_k$, the
direction of unravelling the nesting levels, and whether an additional
linearization step is applied. Typical choices for $E_k$ are given in column
$E_k$, which can be based on a penalizer of the forward model
$z_{k+1}=f_k(z_k;\theta_k)$, using convex (Fenchel) conjugates,
$G_{k+1}(z_{k+1})-z_{k+1}\tr W_kz_k$, or proximal mappings,
$\tfrac{1}{2}\Norm{z_{k+1}-W_kz_k}^2 + F_{k+1}(z_{k+1})$. The last column
indicates the difficulty of the resulting inference task needed to determine
the network activations/targets. The different realizations have also an
impact on the required level of synchronization points: BP and PCN use a
dedicated forward and backward pass (which is relaxed in~\citet{song2020can}),
whereas CHL requires two global (free and clamped) but non-directed phases for
learning. Lifted neural networks such as the method of auxiliary coordinates
(MAC)~\citep{carreira2014distributed}
and lifted proximal
operator machines (LPOM)~\citep{li2018lifted} (and the closely related lifted
Fenchel networks~\citep{gu2018fenchel}) need local synchronization between
layers due to the non-convexity of the underlying objective function. Fenchel
back-propagation is an algorithm similar to back-propagation and is presented
in Section~\ref{sec:FenchelBP}.

Not all instances of inference learning methods
belong to this taxonomy. One aspect that
the above methods have in common is, that they include back-propagation as a
limit case, which is violated e.g.\ for the inference learning methods
proposed in~\citep{zhang2017convergent,askari2018lifted} (as demonstrated
in~\citet{zach2019contrastive}).

Our main contribution is establishing connections between certain
reformulations for deeply nested optimization tasks and inference learning
methods. Singly nested i.e.\ bilevel optimization has been leveraged in the
machine learning literature for parameter learning in variational and
dictionary-based models
(e.g.~\citet{mairal2011task,kunisch2013bilevel,zhou2016bilevel}) and for
meta-learning (such
as~\citet{bennett2008bilevel,franceschi2018bilevel,rajeswaran2019meta,lorraine2020optimizing}). The
resulting optimization methods are often based on surrogates for the original
bilevel optimization tasks using KKT
conditions~\citep{bennett2008bilevel,kunisch2013bilevel,zhou2016bilevel}. Unrolling
a minimization problem is a common technique in machine learning to replace
the inner optimization task with a finite sequence of local minimization
steps~\citep{domke2012generic,maclaurin2015gradient,ochs2016techniques,franceschi2018bilevel}
(which is similar but not identical to back-propagation through
time~\citep{werbos1990backpropagation}). If the inner problem is twice
differentiable, implicit differentiation is applicable for gradient-based
optimization of the outer
objective~\citep{pineda1987generalization,almeida1990learning,kolstad1990derivative,bengio2000gradient,scarselli2008graph},
which has been more recently extended to allow constraints in the inner
problem~\citep{gould2019deep}.

\section{Surrogates for Bilevel Programs}
\label{sec:bilevel}

Since the learning problem in deep networks in Eq.~\ref{eq:DNN} is a nested
minimization task, we first consider singly nested i.e.\ bilevel
problems. For an outer loss $\ell(\cdot)$ and inner objective
$E(\cdot;\cdot)$ we introduce the bilevel minimization problem
\begin{align}
  \label{eq:bilevel}
  \cL(\theta) := \ell(z^*) \qquad \text{s.t. } z^* = \arg\min\nolimits_z E(z; \theta),
\end{align}
where we assume that $E$ has a unique minimizer for each feasible value
$\theta$. We also require that the outer loss $\ell$ is differentiable, bounded from below and
that $\ell(z) < \infty$ for all $z$.
The first step is to rewrite Eq.~\ref{eq:bilevel} using an \emph{optimal value
  reformulation} for bilevel programs~\citep{outrata1988note} (and
see~\citet{dempe2013bilevel} for a discussion of several reformulations for
bilevel programs),
\begin{align}
  \min\nolimits_{\theta,z} \ell(z) \qquad \text{s.t. } E(z;\theta) \le \min\nolimits_{z'} E(z';\theta),
  \label{eq:reformulation}
\end{align}
The corresponding Lagrangian
relaxation is (using $\beta^{-1}$ to parametrize the non-negative multiplier),
\begin{align}
  \max\nolimits_{\beta>0} \min\nolimits_{\theta,z} \ell(z) + \tfrac{1}{\beta} \big( E(z;\theta) -  \min\nolimits_{z'} E(z';\theta) \big).
\end{align}
By fixing a multiplier $\beta>0$ we arrive at the main quantity of interest in this work:
\begin{definition}
  \label{def:surrogates}
  The \emph{contrastive surrogate} for the bilevel minimization problem in
  Eq.~\ref{eq:bilevel} is defined as follows,
  \begin{align}
    \cL^\beta(\theta) &:= \min_z \big\{ \ell(z) + \tfrac{1}{\beta} E(z;\theta) \big\} - \min_z \tfrac{1}{\beta} E(z; \theta) \nonumber \\
    &= \ell(\hat z(\beta)) + \tfrac{1}{\beta} E(\hat z(\beta);\theta) - \tfrac{1}{\beta} E(z^*;\theta),
    \label{eq:surrogate}
  \end{align}
  where $\beta>0$ is the ``spacing parameter.'' The minimizers of the
  two sub-problems are denoted as
  \begin{align}
    \hat z(\beta) & := \arg\min\nolimits_z \beta\ell(z) + E(z;\theta) \nonumber \\
    z^* &:= \arg\min\nolimits_z E(z;\theta),
  \end{align}
  where the dependence on $\theta$ is implicit.
\end{definition}
For brevity we will often write $\hat z_\beta$ instead of $\hat z(\beta)$.
We call this surrogate contrastive in analogy with e.g.\ contrastive Hebbian
learning~\citep{movellan1991contrastive,xie2003equivalence} that relies on
solving two minimization tasks; one to obtain a loss-augmented ``clamped''
solution and one for the ``free'' solution not using any information from the
outer loss~$\ell$.
The contrastive surrogate is a ``min-min-max'' instance, as one aims to
minimize w.r.t.\ $\theta$, but the two subproblems in $z$ have opposing
signs. Fortunately, in our setting solving $\min_z E(z;\theta)$ is typically
(but not always) easy and often has a closed-form solution. If
$E(z;\theta)$ is convex in $z$, then one can obtain a ``min-min-min'' problem
using duality~\citep{zach2020truncated}.

Naturally, we are most interested in the setting when $\beta\approx 0$,
where the contrastive surrogates are good approximations of
the optimal value reformulation (Eq.~\ref{eq:reformulation}), and it is
therefore important to better understand the finite difference
$\tfrac{1}{\beta}( E(\hat z_\beta;\theta) - E(z^*;\theta) )$ in the limit
$\beta\to 0^+$, i.e.\
$\frac{dE(\hat z_\beta;\theta)}{d\beta}|_{\beta=0^+}$. If $E$ is
differentiable w.r.t.\ $z$, then (by the chain rule) the relevant quantity to
analyze is $\frac{d\hat z_\beta}{d\beta}|_{\beta=0^+}=\frac{d\hat z_\beta}{d\beta}|_{\beta=0}$ (due to strong differentiability).
The simplest setting is when
$\ell$ and $E$ are sufficiently regular:
\begin{proposition}
  \label{prop:implicit_diff}
  If $\ell$ is differentiable and $E$ is twice continuously differentiable, then
  \begin{align}
    \tfrac{d}{d\beta} \hat z(\beta)\big|_{\beta=0}
    = -\big( \tfrac{\partial^2}{\partial z^2} E(z^*;\theta) \big)^{-1} \ell'(z^*)
  \end{align}
  and
  \begin{align}
    \tfrac{d}{d\beta} E(\hat z(\beta);\theta)\big|_{\beta=0}
    \!=\! -\tfrac{\partial}{\partial z} E(z^*;\theta) \big( \tfrac{\partial^2}{\partial z^2} E(z^*;\theta) \big)^{-1} \ell'(z^*).
  \end{align}
\end{proposition}
The proof is based on implicit differentiation and is detailed (together with
all following results) in the appendix.
By taking the partial derivative w.r.t.\ $\theta$, we arrive at a familiar
expression for gradient back-propagation through
minimization~\citep{pineda1987generalization,almeida1990learning,kolstad1990derivative,bengio2000gradient,scarselli2008graph},
\begin{align}
  \tfrac{d}{d\theta} \cL(\theta)
  &= -\!\tfrac{\partial^2}{\partial \theta \partial z} E(z^*;\theta) \left( \tfrac{\partial^2}{\partial z^2} E(z^*(\theta);\theta) \right)^{-1} \ell'(z^*)
  \nonumber \\
  &= \lim_{\beta\to 0^+} \tfrac{1}{\beta} \tfrac{\partial}{\partial\theta} \big( E(\hat z_\beta;\theta) - E(z^*;\theta) \big).
  \label{eq:grad_conv}
\end{align}
Hence, we may facilitate gradient-based minimization of $\theta$ e.g.\ by
choosing $\beta\approx 0$ and updating $\theta$ according to
\begin{align}
  \theta^{(t+1)} \gets \theta^{(t)} - \tfrac{\eta^{(t)}}{\beta} \tfrac{\partial}{\partial\theta} \big( E(\hat z_\beta;\theta) - E(z^*;\theta) \big),
  \label{eq:theta_update}
\end{align}
where $\eta^{(t)}$ is the learning rate at iteration $t$. With the assumptions
given in Prop.~\ref{prop:implicit_diff} we know that in the limit
$\beta\to 0^+$ we obtain an exact gradient method. Implicit differentiation
can be extended to allow (smooth) constraints on $z$ in
$E(z;\theta)$~\citep{gould2019deep}. One advantage of the surrogates in
Def.~\ref{def:surrogates} over implicit differentiation is, that the former
requires only the directional derivative
$\frac{d\hat z(\beta)}{d\beta}|_{\beta=0^+}$
to exist, which is a weaker
requirement than existence of a strong derivative
$dz^*(\theta)/d\theta$ as it can be seen in the following simple, but
important example.

\begin{example}[ReLU activation]
  \label{ex:ReLU}
  Let $E(z;\theta)$ be given as
  $E(z;\theta) = \tfrac{1}{2}\norm{z-\theta x}^2 + \imath_{\ge 0}(z)$ and
  $\ell(z) = g\tr z$. Then $z^* = [\theta x]_+$ and
  $\hat z_\beta = [\theta x - \beta g]_+$, and therefore
  \begin{align}
    \frac{d\hat z_\beta}{d\beta}\big|_{\beta=0^+}
    &= \lim_{\beta\to 0^+} \frac{[\theta x - \beta g]_+ - [\theta x]_+}{\beta} = \m D\, g,
  \end{align}
  where $\m D$ is a diagonal matrix with
  \begin{align}
    \m D_{jj} = \begin{cases}
      {-1} & \text{if } (\theta x)_j > 0 \lor \big( (\theta x)_j = 0 \land g_j \ge 0\big) \\
      0 & \text{if } (\theta x)_j < 0 \lor \big( (\theta x)_j = 0 \land g_j < 0 \big).
    \end{cases}
  \end{align}
  Thus, the gradient w.r.t.\ $\theta$ is given by
  \begin{align}
    \tfrac{\partial }{\partial\theta} \lim\nolimits_{\beta\to 0^+} \cL(\beta) = \tfrac{d\hat z_\beta}{d\beta}\big|_{\beta=0^+} x\tr = \m D\,g\, x\tr.
  \end{align}
  This derivative always exists, which is in contrast to back-propagation and
  implicit differentiation~\citep{gould2019deep}, which have both undefined
  gradients at $(\theta x)_j=0$.
\end{example}

This example can be extended to the following result:
\begin{proposition}
  \label{prop:z_derivative}
  Let
  \begin{align}
    \label{eq:constrained_program}
    E(z;\theta) = f(z;\theta) \qquad \text{s.t. } g_i(z) \ge 0,
  \end{align}
  where $f$ and every $g_i$ are twice continuously differentiable w.r.t.\
  $z$. We assume all constraints are active at the solution
  $z^*=\arg\min_{z} E(z;\theta)$, i.e.\ $g_i(z^*)=0$ for all $i$, and ignore
  inactive constraints with $g_i(z^*)>0$. We also assume that
  $\tfrac{\partial^2 f(z^*;\theta)}{\partial z^2}$ is (strictly) positive
  definite and that a suitable constraint qualification holds (such as
  LICQ). Let $\mathcal W^*$ be the indices of weakly active contraints at
  $z^*$, and
  \begin{align}
    \v c &:= \ell'(z^*)
    & \m A &:= \tfrac{\partial^2}{\partial z^2} f(z^*;\theta) \!+\! \sum\nolimits_i \lambda_i^* \tfrac{\partial^2}{\partial z^2} g_i(z^*)
    & \m B &:= \tfrac{\partial}{\partial z} \v g(z^*).
  \end{align}
  If $\mA$ is (strictly) p.d., then
  \begin{align}
    \tfrac{d}{d\beta} \hat z(\beta) \big|_{\beta=0^+} = \dot z,
  \end{align}
  where $\dot z$ is the unique solution of the following strictly convex
  quadratic program,
  \begin{align}
    \label{eq:QP_z_deriv}
    \tfrac{1}{2} \dot z\tr \mA \dot z + \v c\tr \dot z \;\;\; \text{s.t. } \mB \dot z \ge 0
    \;\;\; \forall i\notin \mathcal W^*: \mB_{i,:} \dot z = 0.
  \end{align}
\end{proposition}

\begin{remark}
  If we reparametrize $\dot u = \mA^{1/2}\dot z$, then (with
  $\m C := \mB\mA^{-1/2}$) the QP in Eq.~\ref{eq:QP_z_deriv} is equivalent to
  \begin{align}
    \tfrac{1}{2} \norm{\dot u + \mA^{-1/2}\v c}^2 \;\; \text{s.t. } \m C \dot u \ge 0
    \;\; \forall i\notin \mathcal W^*: \m C_{i,:} \dot u = 0,
  \end{align}
  which projects the natural descent direction $-\mA^{-1/2}\v c$ to the
  feasible (convex) region induced by strongly and weakly active constraints.
\end{remark}

Prop.~\ref{prop:z_derivative} is slightly different to Prop.~4.6
in~\citep{gould2019deep}, which is not applicable when constraints are weakly
active. It assigns a sensible derivative e.g.\ for DNN units with ReLU
activations as seen in the above example (at this point this applies only to
networks with a single layer, and we refer to Section~\ref{sec:nested} for a
discussion of deep networks). Hard sigmoid and hard tanh activation functions
are obtained by contraining $z$ to be inside the hyper-cubes $[0,1]^n$ and
$[-1,1]^n$, respectively, where $n$ is the dimension of $z$. If we have only
strongly active constraints, then (using the Woodbury matrix identity)
\begin{align}
  \frac{d\hat z(\beta)}{d\beta}\big|_{\beta=0^+} &= -\lim_{\mu\to\infty} \left( \mA + \mu \mB\tr\mB \right)^{-1} \ell'(z^*) \nonumber \\
  {} &= -\big( \mI - \m A^{-1} \m B\tr ( \m B \m A^{-1} \m B\tr )^{-1} \m B \big) \m A^{-1} \ell'(z^*) \nonumber
\end{align}
as in~\citep{gould2019deep}. When there are many weakly active constraints,
then solving the QP in Eq.~\ref{eq:QP_z_deriv} is not
straightforward. Fortunately, when modeling standard DNN layers, the
subproblems typically decouple over units, and we have at most a single
(strongly or weakly) active constraint (as in the ReLU example above), and
closed-form expressions are often available.
Once $d\hat z(\beta)/d\beta|_{\beta=0^+}$ is determined, gradient-based optimization of
$\theta$ is straightforward:
\begin{corollary}
  In addition to the assumptions in Prop.~\ref{prop:z_derivative}, let
  $f(z,\theta)$ be differentiable in $\theta$. Then
  \begin{align}
    \tfrac{\partial }{\partial\theta} \lim\nolimits_{\beta\to 0^+} \cL^\beta(\theta)
    = \frac{\partial f(z^*;\theta)}{\partial\theta} \cdot \frac{d\hat z(\beta)}{d\beta}\big|_{\beta=0^+}.
  \end{align}
\end{corollary}
This result can be extended if the constraints depend on $\theta$ by
incorporating the multipliers $\lambda^*$.

So far we focused on the limit case $\beta\to 0^+$,
where the contrastive surrogates yield a generalized chain rule. In practice
it is absolutely possible to use non-infinitesimal values for $\beta$ when
optimizing $\cL^\beta$. We obtain under convexity assumptions the following
approximation property:

\begin{proposition}
  \label{prop:approximation}
  Let $\ell$ be convex and $E(\cdot;\theta)$ be strongly convex with parameter
  $\nu>0$ for all $\theta$. Then we obtain
  \begin{align}
    \cL^\beta(\theta) \le \cL(\theta) \le \cL^\beta(\theta) + \tfrac{\beta}{2\nu} \norm{\partial\ell(z^*)}^2
    \label{eq:approximation}
  \end{align}
  for all $\beta>0$, where $\partial\ell(z^*)$ is any subgradient of $\ell$ at $z^*$.
\end{proposition}
This result immediately implies, that if $\ell$ has bounded (sub-)gradients
(such as the $\ell_1$ loss), then $\cL^\beta$ stays within a fixed band near
$\cL$ for all $\theta$.

\section{Surrogates for Deeply Nested Minimization}
\label{sec:nested}

The previous section has discussed single level surrogates for bilevel
minimization tasks, and in this section we move towards deeply nested
problems. A nested problem (with $L$ nesting levels) is given as follows
(recall Eq.~\ref{eq:DNN}),
\begin{align}
  \label{eq:deepnested}
  \cL_{\mathrm{deep}}(\theta) := \ell(z_L^*) \;\;\text{s.t. } z_1^* &= \arg\min\nolimits_{z_1} \!E_1(z_1; \theta) \\
  z_k^* &= \arg\min\nolimits_{z_k} \!E_k(z_k, z_{k-1}^*; \theta) \nonumber
\end{align}
for $k=2,\dotsc,L$.
We assume that each arg-min is unique and can be stated as a function,
\begin{align}
  z_k^* = \arg\min\nolimits_{z_k} E_k(z_k, z_{k-1}; \theta) = f_k(z_{k-1}; \theta) \nonumber.
\end{align}
This is the case e.g.\ when $E_k$ is strictly convex in $z_k$.
In order to avoid always handling $E_1$ specially, we formally introduce
$z_0$ as argument to $E_1=E_1(z_1,z_0;\theta)$. $z_0$ could be a purely dummy
argument or---in the case of DNNs---the input signal provided to the network.

Gradient-based optimization of Eq.~\ref{eq:deepnested} is usually relying on
back-propagation, which can be seen as instance of the chain
rule~(e.g.~\citet{rumelhart1986learning}) or as the adjoint equations from an
optimal control perspective~\citep{lecun1988theoretical}. In contrast to these
approaches, we proceed in this and in the next section by iteratively applying
the contrastive surrogates introduced in Section~\ref{sec:bilevel}.
We have two natural ways to apply the contrastive surrogates to deeply nested
problems: either top-down (outside-in) or bottom-up (inside-out), which are discussed separately below.

\subsection{Bottom-Up Expansion}
\label{sec:bottom_up}

The first option is to expand recursively from the innermost (bottom) nesting
level, i.e.\ replacing the subproblem
\begin{align}
  \min\nolimits_{z_2} E_2(z_2,z_1^*;\theta) \qquad \text{s.t. } z_1^* = \arg\min\nolimits_{z_1} E_1(z_1;\theta) \nonumber
\end{align}
with
\begin{align}
  \min_{z_1,z_2} \left\{ E_2(z_2,z_1;\theta) + \tfrac{1}{\beta_1} E_1(z_1;\theta) \right\} - \min_{z_1} \tfrac{1}{\beta_1} E_1(z_1;\theta), \nonumber
\end{align}
which yields a problem with $L-1$ levels. Continuing this process yields the
following proposition:
\begin{proposition}
  \label{prop:bottom_up}
  Let $\bbeta=(\beta_1,\dotsc,\beta_L)$ be a vector of positive spacing
  parameters. Applying the contrastive surrogate in Def.~\ref{def:surrogates}
  recursively on Eq.~\ref{eq:deepnested} from the innermost to the outermost
  level yields the following \emph{global contrastive} objective,
  \begin{align}
    \label{eq:GCL}
    \cL^{\bbeta}_{GC}(\theta) := \min_{z_1,\dotsc,z_L} \left\{ \ell(z_L) + \sum\nolimits_{k=1}^{L} \tfrac{E_k(z_k,z_{k-1};\theta)}{\prod_{l=k}^{L}\beta_l} \right\}
    - \min_{z_1,\dotsc,z_L} \left\{ \sum\nolimits_{k=1}^{L} \tfrac{E_k(z_k,z_{k-1}; \theta)}{\prod_{l=k}^{L}\beta_l} \right\}.
  \end{align}
\end{proposition}
We also introduce the free ``energy'' $\cE^{\bbeta}$ and its loss-augmented
(``clamped'') counterpart $\hat\cE^{\bbeta}$,
\begin{align}
  \cE^{\bbeta}(z;\theta) &:= \sum\nolimits_{k=1}^{L} \tfrac{1}{\prod_{l=k}^{L}\beta_l} E_k(z_k,z_{k-1};\theta) \nonumber \\
  \hat\cE^{\bbeta}(z;\theta) &:= \ell(z_L) + \cE^{\bbeta}(z;\theta).
\end{align}
Hence, $\cL^{\bbeta}(\theta)$ is the difference of the optimal clamped
(loss-augmented) and free energies,
\begin{align}
  \cL_{GC}^{\bbeta}(\theta) &= \min_z \hat\cE^{\bbeta}(z;\theta) - \min_z \cE^{\bbeta}(z;\theta) \nonumber \\
  {} &=  \min_{\hat z} \max_{\check z} \hat\cE^{\bbeta}(\hat z;\theta) - \cE^{\bbeta}(\check z;\theta).
\end{align}
The minimizers $\check z$ and $\hat z$ are called the \emph{free} and
\emph{clamped} solution, respectively. We call $\cL_{GC}^{\bbeta}$ a globally
contrastive loss, since it is the difference of two objectives involving all
(nesting) layers. This bottom-up expansion requires (approximately) solving
two minimization tasks to obtain the free and the clamped solution in order to
form a gradient w.r.t.\ $\theta$. In analogy to learning in Boltzmann
machines, these two minimization problems are referred to as \emph{free phase}
and the \emph{clamped
  phase}~\citep{movellan1991contrastive,xie2003equivalence}. When
$\cE^{\bbeta}$ is jointly convex in $z$ (which is sufficient to model general
function approximation such as ReLU-based DNNs), the ``min-max'' problem
structure can be converted to a ``min-min'' one using convex
duality~\citep{zach2019contrastive,zach2020truncated}. By construction we also
have $\cE^{\bbeta}(\hat z;\theta) \ge \cE^{\bbeta}(\check z;\theta)$, and if
$\cE^{\bbeta}$ is strictly convex, then
$\cE^{\bbeta}(\hat z;\theta) = \cE^{\bbeta}(\check z;\theta)$ iff
$\hat z=\check z$. Hence, learning in this framework is achieved by reducing
the gap between a clamped (loss-augmented) and a free energy that cover all
layers.
If we set all $\beta_k$ identical to a common feedback parameter
$\beta>0$, then the free energy $\cE^{\bbeta}$ reduces to
\begin{align}
  \cE^{\bbeta}(z;\theta) &:= \sum\nolimits_{k=1}^{L} \beta^{k-1-L} E_k(z_k,z_{k-1}; \theta).
\end{align}
Typically $\beta\in(0,1)$ and therefore later layers are discounted. Such
contrastive loss between discounted terms is referred as \emph{contrastive
  Hebbian learning}, and their relation to classical back-propagation for
specific instances of $E_k$ is analysed
in~\citep{xie2003equivalence,zach2019contrastive}.

One advantage of the global contrastive framework over the approach described
in the next section is, that the free and the clamped energies,
$\cE^{\bbeta}$ and $\hat\cE^{\bbeta}$, can be chosen to be jointly convex in
all activations $z_1,\dotsc,z_L$ while still approximating e.g.\ general
purpose DNNs with ReLU activation functions to arbitrary precision. By
modifying the terms $E_k$ connecting adjacent layers, one can also obtain
Lipschitz continuous deep networks by design, that exhibit better robustness
to adversarial perturbations~\citep{hoier2020LRRN}.

\subsection{Top-Down Expansion}
\label{sec:top_down}

Applying the analogous procedure described in Section~\ref{sec:bottom_up} from
the outermost (top) level successively to the inner nesting levels yields a
different surrogate for deeply nested problems:
\begin{proposition}
  \label{prop:top_down}
  Let $\bbeta = (\beta_1,\dotsc,\beta_L)$ with each $\beta_k>0$ and let
  $\check E_k$ and $\tilde E_k$ be given by
  \begin{align}
    \check E_k(z_{k-1};\theta) &:= \min\nolimits_{z_k'} E_k(z_k',z_{k-1}; \theta) \\
    \tilde E_k(z_k, z_{k-1};\theta) &:= E_k(z_k, z_{k-1};\theta) - \check E_k(z_{k-1};\theta)
  \end{align}
  for each $k=1,\dotsc,L$. Applying the expansion in Def.~\ref{def:surrogates}
  recursively on Eq.~\ref{eq:deepnested} from the outermost to the innermost
  level yields the following \emph{local (or layer-wise) contrastive}
  objective,
  \begin{align}
    \cL_{\mathrm{LC}}^{\bbeta}(\theta) &= \!\! \min_{z_1,\dotsc,z_L} \Big\{ \ell(z_L)
    + \sum\nolimits_{k=1}^L \tfrac{1}{\beta_k} \tilde E_k(z_k, z_{k-1};\theta) \Big\}. \label{eq:LCL}
  \end{align}
  We also introduce
  \begin{align}
    \tilde\cE^{\bbeta}(z;\theta) := \sum\nolimits_{k=1}^L \tfrac{1}{\beta_k} \tilde E_k(z_k, z_{k-1};\theta).
  \end{align}
\end{proposition}
Similar to the global contrastive setting,
$\tilde\cE^{\bbeta}(z;\theta)\ge 0$ for all $z$; and if
$E_k(z_k, z_{k-1};\theta)$ is strictly convex, then
$\tilde\cE^{\bbeta}(z;\theta)=0$ iff
$z_k=\arg\min_{z_k'} E_k(z_k',z_{k-1})$ for all $k=1,\dotsc,L$. Learning the
parameters $\theta$ aims to reduce the discrepancy measure
$\tilde\cE^{\bbeta}$. In contrast to the global contrastive framework, the
learning task is a pure min-min instance, and it requires only maintaining one
set of network activations $z$. One downside of the local contrastive method
is, that $\tilde\cE^{\bbeta}$ will be difficult to optimize in any interesting
setting (i.e.\ with a non-linear activation function), even if $\ell$ and all
$E_k$ are convex. This is due to the following reasoning:
$\tilde\cE^{\bbeta}(z;\theta)=0$ iff $z_k = f_{k-1}(z_{k-1})$, which is a
non-linear constraint in general (unless $f_{k-1}$ is linear). Hence, with at
least one $f_k$ being nonlinear, the set
$\{z:z_k = f_{k-1}(z_{k-1}) \forall k\}$ is non-convex, and therefore
$\tilde\cE^{\bbeta}$ cannot be jointly convex in $z_1,\dotsc,z_L$. While joint
convexity is out of reach, it is possible to obtain instances that are
layer-wise convex (cf.~Section~\ref{sec:LPOM}).

\subsubsection{Auxiliary coordinates and predictive coding networks}

Starting from very different motivations, both predictive coding
networks~\citep{whittington2017approximation,song2020can} and the method of
auxiliary coordinates~\citep{carreira2014distributed} arrive at the same
underlying objective,
\begin{align}
  \cE_{\text{MAC}}(z;\theta) = \ell(z_K) +\! \sum_k \tfrac{\mu_k}{2} \norm{z_k \!-\! f_{k-1}(z_{k-1}; \theta)}^2,
\end{align}
which is essentially a quadratic penalizer for the (typically non-linear and
non-convex) constraint $z_k = f_{k-1}(z_{k-1}; \theta)$. By identifying
$\mu_k=1/\beta_k$ and by using
$E_k(z_k,z_{k-1};\theta) = \tfrac{1}{2}\Norm{z_k-f_{k-1}(z_{k-1}; \theta)}^2$,
we identify $\cE_{\text{MAC}}$ as an instance of a local contrastive objective
(Eq.~\ref{eq:LCL}). At the same time it can be also interpreted as an instance
of the global contrastive objective (Eq.~\ref{eq:GCL}), after equating
$\mu_k^{-1} = \prod_{l=k}^L\beta_l$, since the optimal free energy satisfies
$\min_z \cE^{\bbeta}(z;\theta)=0$. The two interpretations differ in the
meaning of the multipliers $\mu_k$, depending on whether they encode
discounting of later layers. In contrast to lifted proximal operator machines
discussed in the next section (Section~\ref{sec:LPOM}), inferring the
activations $(z_k)_{k=1}^L$ is usually not even layerwise convex.

\subsubsection{Lifted Proximal Operator Machines}
\label{sec:LPOM}

A different choice for $E_k$ is inspired by convex conjugates and the
Fenchel-Young inequality: let $G_k(\cdot;\theta)$ be l.s.c.\ and convex for
$k=1,\dotsc,L$, and define $E_k$ as follows,
\begin{align}
  \label{eq:LPOM_Ek}
  E_k(z_k,z_{k-1};\theta) := G_k(z_k;\theta) - z_k\tr W_{k-1} z_{k-1},
\end{align}
where the trainable parameters $\theta=(W_0,\dotsc,W_{L-1})$ are the weight
matrices.
Recall $\check E_k(z_{k-1};\theta) = \min_{z_k} E_k(z_k,z_{k-1};\theta)$ (Prop.~\ref{prop:top_down}), and therefore
we obtain
\begin{align}
  \check E_k(z_{k-1};\theta) &= \min\nolimits_{z_k} E_k(z_k,z_{k-1};\theta) \nonumber \\
  &= \min\nolimits_{z_k} G_k(z_k;\theta) - z_k\tr W_{k-1}z_{k-1} = -\max_{z_k} z_k\tr W_{k-1}z_{k-1} G_k(z_k;\theta) \nonumber \\
  &= -G_k^*(W_{k-1}z_{k-1}),
\end{align}
where $G_k^*$ is the convex conjugate of $G_k$. Thus, in this setting we read
\begin{align}
  \tilde E_k(z_k,z_{k-1}) = E_k(z_k,z_{k-1})-\check E_k(z_{k-1}) = G_k(z_k) - z_k\tr W_{k-1}z_{k-1} + G_k^*(W_{k-1}z_{k-1}),
\end{align}
hence $\cL_{\text{LC}}^{\bbeta}$ specializes to the following \emph{lifted
  proximal operator machine} objective,
\begin{align}
  \cL_{\text{LPOM}}^{\bbeta}(\theta) = \min_{z_1,\dotsc,z_L} \ell(z_L) +
  \sum\nolimits_{k=1}^L \tfrac{1}{\beta_k} \big( G_k(z_k) - z_k\tr W_{k-1}z_{k-1} \!+\! G_k^*(W_{k-1}z_{k-1})  \big).
  \label{eq:LPOM}
\end{align}
By collecting terms dependent only on $z_k$, minimizing over $z_k$ (for fixed
$z_{k-1}$ and $z_{k+1}$) amounts to solving the following convex minimization problem,
\begin{align}
  \min\nolimits_{z_k} \tfrac{1}{\beta_k} \big( G_k(z_k) - z_k\tr W_{k-1}z_{k-1} \big)
  + \tfrac{1}{\beta_{k+1}} \big( G_{k+1}^*(W_kz_k) - z_k\tr W_k\tr z_{k+1} \big),
\end{align}
hence the minimization task in Eq.~\ref{eq:LPOM} is block-convex w.r.t.\
$z_1,\dotsc,z_L$.
The above choice for $E_k$ is a slight generalization of lifted proximal
operator machines proposed by~\citet{li2018lifted}, where a similar
block-convex cost is proposed for DNNs with element-wise activation functions
$f_k$. In the notation of~\citet{li2018lifted}, $G_k$ corresponds to
$\tilde f_k$ and its convex conjugate $G_k^*$ is represented by
$\tilde g_k$. The ability to utilize general convex $G_k$ e.g.\ allows to
model soft arg-max layers (often referred to as just soft-max layers) as
LPOM-type objective by using the convex conjugate pair
\begin{align}
  G_k(z_k) &= \sum\nolimits_j z_{k,j} \log z_{k,j} & G_k^*(z_{k-1}) &= \log \sum\nolimits_j e^{z_{k-1,j}}. \nonumber
\end{align}
This activation function cannot be represented in the original LPOM framework,
since it requires coupling between the elements in $z_{k-1}$.

\section{Fenchel back-propagation}
\label{sec:fenchel_bp}

First, we observe that Props.~\ref{prop:implicit_diff}
and~\ref{prop:z_derivative} also hold if the outer loss $\ell$ in the
respective reformulation is replaced by its first-order Taylor approximation at~$z^*=\arg\min_z E(z;\theta)$,
\begin{align}
  \bar\ell(z;z^*) := \ell(z^*) + (z-z^*)\tr \ell'(z^*),
\end{align}
since these results only make use of the derivative $\ell'(z^*)$. Hence, in
this section we consider to replace $\ell$ by its linear approximation
at~$z^*$:
\begin{definition}
  For a differentiable outer loss $\ell$, an inner objective $E$ and
  $\beta>0$ the \emph{linearized contrastive surrogate} is given by
  \begin{align}
    \bar\cL^\beta(\theta) := \min_z \big\{ \bar\ell(z;z^*) + \tfrac{1}{\beta} E(z;\theta) \big\} - \min_z \tfrac{1}{\beta} E(z; \theta).
    \label{eq:lin_surrogate}
  \end{align}
\end{definition}
Props.~\ref{prop:implicit_diff} and~\ref{prop:z_derivative} remain valid with
$\cL^\beta$ replaced by $\bar\cL^\beta$. It is now natural to ask what happens
if we use $\bar\cL^\beta$ as building block to convert deeply nested programs,
Eq.~\ref{eq:deepnested}. It can be shown (see the appendix)
that the bottom-up traversal of the nesting levels (corresponding to
$\cL^{\bbeta}_{\mathrm{GC}}$) yields non-promising reformulations and are
therefore of no further interest. Thus, we focus on the top-down approach below.

In order to apply the linearized contrastive surrogate
(Eq.~\ref{eq:lin_surrogate}) in the top-down direction, we need the following
result in order to apply the surrogate recursively when the outer loss is
itself a minimization task:
\begin{proposition}
  \label{prop:lin_top_down_step}
  Let a bilevel problem be of the form
  \begin{align}
    \min\nolimits_y h(y,z^*) \qquad \text{s.t. } z^* = \arg\min\nolimits_z E(z;\theta),
  \end{align}
  where $h(y,z)$ is differentiable w.r.t.\ $z$ for all $y$. Then for a chosen
  $\beta>0$ the linearized contrastive reformulation is given by
  \begin{align}
    \min\nolimits_z & \left\{ \bar h(y^*, z; z^*) + \tfrac{1}{\beta} E(z;\theta) \right\} - \tfrac{1}{\beta} \min\nolimits_z E(z;\theta),
  \end{align}
  where $y^* := \arg\min_y h(y,z^*)$ and
  \begin{align}
    \bar h(y, z; z^*) := h(y,z^*) + (z-z^*)\tr \tfrac{\partial}{\partial z} h(y, z^*).
  \end{align}
\end{proposition}
With this proposition we are able to apply the linearized contrastive
surrogate in top-down direction:
\begin{proposition}
  \label{prop:lin_top_down}
  Let $\bbeta=(\beta_1,\dotsc,\beta_L)$ with each $\beta_k>0$. We recursively
  define the following quantities:
  \begin{align}
    z_1^* &:= \arg\min\nolimits_{z_1} E_1(z_1;\theta) & z_k^* &:= \arg\min\nolimits_{z_k} E_k(z_k; z_{k-1}^*)
  \end{align}
  and
  \begin{align}
    \tilde E_{k}(z_{k},z_{k-1}) &:= E_{k}(z_{k},z_{k-1}) - \min\nolimits_{z_{k}'} E_{k}(z_{k}',z_{k-1}) \nonumber \\
    \bar E_k(z_{k},z_{k-1};z_{k-1}^*) &:= \tilde E_k(z_k,z_{k-1})
                                        + (z_{k-1}-z_{k-1}^*)\tr \tfrac{\partial}{\partial z_{k-1}} \tilde E_{k}(z_{k}, z_{k-1}^*)
  \end{align}
  for $k=2,\dotsc,L$. We also introduce for $k=1,\dotsc,L-1$,
  \begin{align}
    \bar z_L &:= \arg\min_{z_L }\bar\ell(z_L; z_L^*) + \tfrac{1}{\beta_L} E_L(z_L,z_{L-1}^*) \nonumber \\
    \bar z_k &:= \arg\min_{z_k} \tfrac{1}{\beta_{k+1}} \bar E_{k+1}(\bar z_{k+1},z_k; z_k^*) + \tfrac{1}{\beta_k} E_k(z_k, z_{k-1}^*).
  \end{align}
  Then the \emph{linearized local contrastive} surrogate
  $\bar\cL^{\bbeta}_{LC}$ for the deeply nested program Eq.~\ref{eq:deepnested}
  is given by
  \begin{align}
    \bar\cL^{\bbeta}_{\text{LC}}(\theta) = \bar\ell(\bar z_L;z_L^*) + \sum_{k=1}^{L} \tfrac{1}{\beta_k} \bar E_k(\bar z_k, \bar z_{k-1}; z_{k-1}^*).
  \end{align}
\end{proposition}
Observe that in $\bar\cL^{\bbeta}_{LC}$ is given in closed-form and does not
require further inference (i.e.\ minimization) w.r.t.\
$z_1,\dotsc,z_L$. The quantities $z_1^*,\dotsc,z_L^*$ are determined in a
forward pass, where as $\bar z_L, \dotsc,\bar z_1$ are obtained in a
respective backward pass by solving linearly perturbed forward problems.

\subsection{Standard Error Back-propagation}

In view of the forward and backward pass outlined in
Prop.~\ref{prop:lin_top_down}, it should not come as a surprise that standard
error back-propagation can be obtained with a suitable choice for
$E_k$. In particular, if we use
$E_{k}(z_{k},z_{k-1}) = \tfrac{1}{2} \Norm{z_{k}-f_{k-1}(z_{k-1})}^2$, where
$f_{k-1}$ is the desired forward mapping between layers $k-1$ and $k$, then we
obtain a family of back-propagation methods. Observe that with this choice we
have $E_{k}(z_{k}^*,z_{k-1}^*)=0$, and
\begin{align}
  \tfrac{\partial}{\partial z_k} \tilde E_{k+1}(\bar z_{k+1}, z_k^*) = f_k'(z_k^*)\tr (f_k(z_k^*) - \bar z_{k+1}) = f_k'(z_k^*)\tr (z_{k+1}^* - \bar z_{k+1})
\end{align}
and therefore
\begin{align}
  \bar z_L = z_L^* - \beta_L \ell'(z_L^*) & & \bar z_k = z_k^* - \tfrac{\beta_k}{\beta_{k+1}} f_k'(z_k^*)\tr (z_{k+1}^* \!-\! \bar z_{k+1}).
\end{align}
We introduce $\eps_k:= \bar z_k-z_k^*$ to obtain the relations
\begin{align}
  \eps_L = -\beta_L \ell'(z_L^*) & & \eps_k = \tfrac{\beta_k}{\beta_{k+1}} f_k'(z_k^*)\tr \eps_{k+1}.
\end{align}
The recursion for $\eps_k$ is essentially back-propagation (and equivalent to
the proposed backward pass in predictive coding
networks~\citep{millidge2020predictive}). Standard back-propagation is
obtained if $\beta_1=\ldots=\beta_L$, otherwise one obtains a descent (but not
necessarily steepest descent) direction.

\subsection{Fenchel Back-Propagation}
\label{sec:FenchelBP}

In analogy to Section~\ref{sec:LPOM} we consider
$E_{k+1}(z_{k+1},z_{k}) = G_{k+1}(z_{k+1}) - z_{k+1}\tr W_{k}z_{k}$, where
$G_{k+1}$ is a strictly convex function for each $k$. $G_{k+1}$ is chosen such
that $\arg\min_{z_{k+1}} E_{k+1}(z_{k+1},z_{k}) = f_{k+1}(W_{k}z_{k})$ for a
desired activation function $f_{k+1}$. We recall from Section~\ref{sec:LPOM} that
\begin{align}
  \tilde E_{k+1}(z_{k+1},z_{k}) = G_{k+1}(z_{k+1}) - z_{k+1}\tr W_{k}z_{k} + G_{k+1}^*(W_{k}z_{k}) \nonumber
\end{align}
(where $G_{k+1}^*$ is the convex conjugate of $G_{k+1}$) and therefore (using
Prop.~\ref{prop:lin_top_down_step})
\begin{align}
  \tfrac{\partial}{\partial z_{k}} \tilde E_{k+1}(z_{k+1},z_{k}^*) = W_{k}\tr \big( \partial G_{k+1}^*(W_{k}z_{k}^*) - z_{k+1} \big)
  = W_{k}\tr \big( f_{k+1}(W_{k}z_{k}^*) - z_{k+1} \big).
\end{align}
Thus, the realization of $\bar\cL^{\bbeta}_{LC}$ in this setting is given by
\begin{align}
  \bar\cL^{\bbeta}_{\mathrm{FenBP}}(\theta) = \bar\ell(\bar z_L;z_L^*) + \sum\nolimits_{k=1}^L \tfrac{1}{\beta_k} \tilde E_k(\bar z_k,z_{k-1}^*)
  + \sum\nolimits_{k=1}^{L-1} \tfrac{1}{\beta_{k+1}} (\bar z_k-z_k^*)\tr W_{k}\tr (z_{k+1}^* - \bar z_{k+1}),
  \label{eq:FenBP}
\end{align}
where FenBP stands for \emph{Fenchel back-propagation}. In contrast to
standard back-propagation, non-infinitesimal target values are propagated
backwards through the layers.  It also combined the power to model
non-differentiable activation functions with the efficiency of
back-propagation.
Specializing this further to ReLU activation functions (using
$G_k(z_k)=\tfrac{1}{2}\norm{z_k}^2 + \imath_{\ge 0}(z_k)$ and thus
$G_k^*(u) = \tfrac{1}{2}\norm{[u]_+}^2$) yields the following relations,
\begin{align}
  \tilde E_k(z_k,z_{k-1}) &= \tfrac{1}{2} \norm{z_k}^2 + \imath_{\ge 0}(z_k) - z_k\tr W_{k-1}z_{k-1} + \tfrac{1}{2} \norm{[W_{k-1}z_{k-1}]_+}^2,
\end{align}
and consequently $z_k^* = [W_{k-1}z_{k-1}^*]_+$,
$z_L^* = W_{L-1}z_{L-1}^*$ for the forward pass, and
\begin{align}
  \bar z_L &= \arg\min_{z_L } z_L\tr \ell'(z_L^*) + \tfrac{1}{\beta_L} \left( \tfrac{1}{2} \norm{z_L}^2 - z_L\tr W_{L-1}z_{L-1}^* \right) \nonumber\\
  &= W_{L-1}z_{l-1}^* - \beta_L \ell'(z_L^*) \nonumber \\
  \bar z_k &= \arg\min_{z_k\ge 0} \tfrac{1}{\beta_{k+1}} z_k\tr W_{k-1}\tr (z_{k+1}^* - \bar z_{k+1})
             + \tfrac{1}{\beta_k} \left( \tfrac{1}{2} \norm{z_k}^2 - z_k\tr W_{k-1}z_{k-1}^* \right) \nonumber \\
  &= \big[ W_{k-1}z_{k-1}^* + \tfrac{\beta_k}{\beta_{k+1}} W_k\tr (\bar z_{k+1} - z_{k+1}^*) \big]_+
\end{align}
determine the backward pass. This means that the forward and backward passes
in Fenchel BP are similar to back-propagation in terms of computation
efficiency, as the only modification are the details of the back-propagated
error signals. As seen in Ex.~\ref{ex:ReLU} the error signal
$\tfrac{1}{\beta_k} (\bar z_k-z_k^*)$ converges to a generalized derivative of
the activation function in the limit case $\beta_k\to 0^+$.

\begin{remark}
  There is a subtle but important issue regarding the correct way of computing
  $\frac{\partial}{\partial W_k} \bar\cL^{\bbeta}_{\mathrm{FenBP}}$. In view
  of Eq.~\ref{eq:FenBP} it is easy to eventually obtain to an incorrect
  gradient
  \begin{align}
    \tfrac{\partial}{\partial W_k} \bar\cL^{\bbeta}_{\mathrm{FenBP}} = \tfrac{1}{\beta_{k+1}} \left( z_{k+1}^* - \bar z_{k+1} \right) \bar z_k\tr,
  \end{align}
  whereas the correct one is given by,
  \begin{align}
    \tfrac{\partial}{\partial W_k} \bar\cL^{\bbeta}_{\mathrm{FenBP}} = \tfrac{1}{\beta_{k+1}} \left( z_{k+1}^* - \bar z_{k+1} \right) (z_k^*)\tr.
    \label{eq:grad_Wk_correct}
  \end{align}
  The reason is, that we have to ignore the linearization terms
  $(\bar z_k-z_k^*)\tr W_{k}\tr (z_{k+1}^* - \bar z_{k+1})$ in
  Eq.~\ref{eq:FenBP}, since they only appear when linearizing
  $\tilde E_k$ with respect to a lower level variable $z_{k-1}$. Using only
  the $\tfrac{1}{\beta_{k+1}} \tilde E_{k+1}(\bar z_{k+1},z_k^*)$ term to
  determine the gradient w.r.t.\ $W_k$ yields the correct answer in
  Eq.~\ref{eq:grad_Wk_correct}. This gradient is the contribution from a
  single training sample and is accumulated accordingly over the training
  set or a respective mini-batch for gradient-based learning.
\end{remark}

\section{Discussion}

\begin{figure*}[tb]
  \centering
  \includegraphics[height=2.15cm]{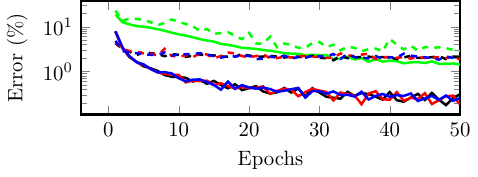}%
  \includegraphics[height=2.15cm]{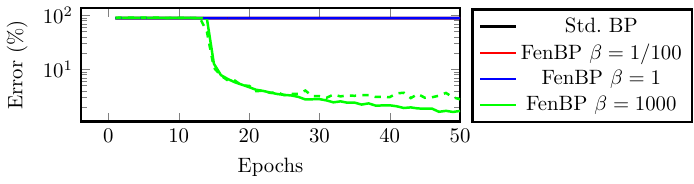}
  \caption{Training progress for 784-256-128-64-10 ReLU MLP trained on
    MNIST. (Left) Uniform Glorot weight initialization, (right) Negative
    weight initialization. Solid curves correspond to training errors (in \%)
    and dashed ones are test errors.}
  \label{fig:errors}
\end{figure*}

In this work we show that several apparently unrelated inference learning
methods can be unified via the framework of contrastive surrogates for deeply
nested optimization problems. By applying Prop.~\ref{prop:z_derivative}
successively on each level we obtain the following corollary as one
consequence of this framework:
\begin{corollary}
  Let $\ell$ be differentiable and $E_k$, $k=1,\dotsc,L$, satisfy the
  assumptions in Prop.~\ref{prop:z_derivative}, then we have
  \begin{align}
    \tfrac{d}{d\theta} \cL_{\mathrm{deep}}(\theta) &= \tfrac{\partial}{\partial\theta} \lim_{\bbeta\to \v 0^+} \cL_{GC}^{\bbeta}(\theta)
    = \tfrac{\partial}{\partial\theta} \lim_{\bbeta\to \v 0^+} \cL_{LC}^{\bbeta}(\theta)
    = \tfrac{\partial}{\partial\theta} \lim_{\bbeta\to \v 0^+} \bar\cL_{LC}^{\bbeta}(\theta).
  \end{align}
\end{corollary}
Thus, the methods discussed in Sections~\ref{sec:nested}
and~\ref{sec:FenchelBP} are equivalent to standard back-propagation in their
respective limit case (and approximate back-propagation for sufficiently small values
$\beta_k>0$). Hence, constructive arguments specific to a proposed method (such
as the ones given in~\citep{xie2003equivalence,zach2019contrastive}) are not
strictly necessary.
Additionally it follows, that back-propagation can be extended to rely solely
on one-sided directional derivatives instead of regular (strong) derivatives
of the activation function.

We conclude by pointing out that in practice the contrastive surrogates are
valid objectives to train a DNN in their own right. While the analysis
conducted in this paper is largely addressing the limit case
$\beta_k\to 0^+$,
when all the discussed
methods converge essentially to back-propagation, even finite values for
$\beta_k>0$
lead to valid learning losses. This is due to $\cL^{\bbeta}_{GC}$,
$\cL_{LC}^{\bbeta}$ and $\bar\cL_{LC}^{\bbeta}$ attaining the minimal possible
objective value only for a perfect predictor (leading also to a minimal target
loss $\ell$).
Empirical results of existing inference learning methods are given in the
respective literature.
At this point we illustrate the numerical behavior of Fenchel BP in
Fig.~\ref{fig:errors} for a 4-layer multi-layer perceptron (MLP) trained on
MNIST using ADAM~\cite{kingma2015adam} as stochastic optimization method
(batch size 50, learning rate $0.001$, uniform Glorot or negative weight
initialization). The training error (solid lines) and the test error (dashed
lines) are depicted using standard and Fenchel back-propagation with different
fixed values for $\beta_k=\beta\in\{0.01,1,1000\}$ for all $k$. It can be seen
in Fig.~\ref{fig:errors}(left), that Fenchel BP for small but non-vanishing
choices of $\beta$ reduces the training error similarly well as standard
back-propagation, and progress is slowed down only for a large value of
$\beta=1000$. On the other hand, a value of $\beta \gg 0$ is able to escape
the vanishing gradient problem (Fig.~\ref{fig:errors}(right)). In particular,
an all-negative weights initialization for the same ReLU-based MLP---which
will not progress using back-propagation due to vanishing gradients---can be
successfully trained by Fenchel BP using a sufficiently large value for
$\beta$ (achieving 1.7\%/2.79\% training and test error compared to
88.66\%/88.65\% for standard BP and Fenchel BP with $\beta\in\{0.01,1\}$).
Understanding and exploring this property and other variations of Fenchel BP
is subject of future work.

\paragraph{Acknowledgements}
We are grateful to Guillaume Bourmaud and to anonymous reviewers for helpful feedback.
This work was partially supported by the Wallenberg AI, Autonomous Systems and Software Program (WASP) funded by the Knut and Alice Wallenberg Foundation.

{
  \small
  \bibliographystyle{plain}
  \bibliography{literature}
}

\appendix

\small

\section{Proofs}

\begin{proof}[Proof of Prop.~\ref{prop:implicit_diff}]
  In the differentiable setting $\hat z(\beta)$ satisfies
  \begin{align}
    0 &= \beta\ell(\hat z(\beta) + \frac{\partial E(\hat z(\beta);\theta)}{\partial z}
  \end{align}
  and total differentiation w.r.t.\ $\beta$ at~0 yields
  \begin{align}
    0 &= \ell'(\hat z(\beta)) + \frac{\partial^2 E(\hat z(\beta);\theta)}{\partial z^2} \frac{d\hat z(\beta)}{d\beta}
        \iff \frac{d\hat z(\beta)}{d\beta} = -\left( \frac{\partial^2 E(\hat z(\beta);\theta)}{\partial z^2} \right)^{-1} \ell'(\hat z(\beta))
  \end{align}
  Letting $\beta\to 0$ and using that $\hat z(0) = z^*$ yields the claimed
  relation,
  \begin{align}
    \frac{d\hat z(\beta)}{d\beta}\big|_{\beta=0} = -\!\left( \frac{\partial^2 E(z^*;\theta)}{\partial z^2} \right)^{-1} \ell'(z^*).
  \end{align}
  The second relation in Prop.~\ref{prop:implicit_diff}
  is obtained by applying the chain rule.
\end{proof}

\begin{proof}[Proof of Prop.~\ref{prop:z_derivative}]
  We restate Eq.~\ref{eq:constrained_program}
  after introducing slack variables $s \ge 0$
  \begin{align}
    E(z;\theta) = f(z;\theta) \qquad \text{s.t. } g_i(z) - s_i = 0 \qquad \forall i\notin\mathcal W^*: s_i=0.
  \end{align}
  The slack variables corresponding to strongly active constraints are forced
  to~0. Now $\hat z(\beta)$ satisfies (where we simply write $\hat z$ instead of $\hat z(\beta)$)
  \begin{align}
    \beta\ell'(\hat z) + f'(\hat z) + \sum\nolimits_i \lambda_i g_i'(\hat z) = 0,
  \end{align}
  where $\lambda_i\in\mathbb{R}$ is the Lagrange multipliers for
  $g_i(z) - s_i$, and
  $\ell'$, $f'$ and $g_i'$ are derivatives w.r.t.\ $z$. Taking the total
  derivative w.r.t.\ $\beta$ yields
  \begin{align}
    \ell'(\hat z) + \beta \ell''(\hat z) \dot z + f''(\hat z) \dot z + \sum\nolimits_i\lambda_i\tr g_i''(\hat z) \dot z + g'(\hat z)\tr \dot\lambda = 0,
  \end{align}
  where $\dot z$ and $\dot\lambda$ are shorthand notations for
  $d\hat z/d\beta$ and $d\lambda/d\beta$. We introduce
  \begin{align}
    \m A := f''(z^*) + \sum\nolimits_i \lambda_i\tr g_i''(z^*) & & \m B := g'(z^*).
  \end{align}
  Thus, the above reads for $\beta \to 0^+$:
  \begin{align}
    \ell'(z^*) + \m A \dot z + \m B\tr \dot\lambda = 0,
  \end{align}
  Taking the derivative of the constraint $g(z) - s = 0$ yields
  $\m B \dot z = \dot s$. If the $i$-th constraint is an equality constraint,
  then $s_i$ is constant~0 and therefore $\dot s_i=0$. Otherwise
  $\dot s_i\ge 0$. Overall, we have a linear system of equations,
  \begin{align}
    \begin{pmatrix} \m A & \m B\tr \\ \m B & \m 0 \end{pmatrix} \binom{\dot z}{\dot\lambda} = \binom{-\ell'}{\dot s}.
  \end{align}
  This system together with $\dot s\ge 0$ corresponds to the KKT conditions of
  the following quadratic program,
  \begin{align}
    \tfrac{1}{2} \dot z\tr \mA \dot z + \v c\tr \dot z \qquad \text{s.t. } \mB \dot z = \dot s \qquad \dot s \ge 0
    \qquad \forall i\notin \mathcal W^*: \dot s_i = 0,
  \end{align}
  which is equivalent to Eq.~\ref{eq:QP_z_deriv}.
\end{proof}

\begin{proof}[Proof of Prop.~\ref{prop:approximation}]
  Recall that
  \begin{align}
    \cL^\beta(\theta) = \min_z \ell(z) + \tfrac{1}{\beta} \big( E(z;\theta) - E(z^*;\theta) \big)
    = \ell(\hat z_\beta) + \tfrac{1}{\beta} \big( E(\hat z_\beta;\theta) - E(z^*;\theta) \big).
  \end{align}
  By using the optimality of $\hat z_\beta$, in particular
  $\ell(\hat z_\beta) + \beta^{-1}E(\hat z_\beta;\theta) \le \ell(z) +
  \beta^{-1}E(z)$ for all $z$, we obtain:
  \begin{align}
    \ell(\hat z_\beta) + \beta^{-1}E(\hat z_\beta;\theta) \le \ell(z^*) + \beta^{-1}E(z^*)
    & \iff \underbrace{\ell(\hat z_\beta) + \beta^{-1} \big( E(\hat z_\beta;\theta) - E(z^*;\theta) \big)}_{=\, \cL^\beta(\theta)}
      \le \underbrace{\ell(z^*)}_{=\, \cL(\theta)} \\
    & \iff \cL^\beta(\theta) \le \cL(\theta),
  \end{align}
  which shows the lower bound. Now let $E(z;\theta)$ be strongly convex in
  $z$ for all $\theta$ with parameter $\nu>0$. Thus,
  \begin{align}
    E(z;\theta) \ge E(z';\theta) + \nabla_z E(z';\theta)\tr (z-z') + \tfrac{\nu}{2} \norm{z-z'}^2
    \implies E(z;\theta) \ge E(z^*;\theta) + \tfrac{\nu}{2} \norm{z-z^*}^2 ,
  \end{align}
  and therefore
  \begin{align}
    \min_z \left\{ \ell(z) - \ell(z^*) + \beta^{-1} E(z;\theta) \right\} &\ge \min_z \left\{ \ell'(z^*)\tr (z-z^*) + \tfrac{1}{\beta} E(z;\theta) \right\} \\
    {} &\ge \min_z \left\{ \ell'(z^*)\tr (z-z^*) + \tfrac{1}{\beta} E(z^*;\theta) + \tfrac{\nu}{2\beta} \norm{z-z^*}^2 \right\} \\
    {} &= \tfrac{1}{\beta} E(z^*) - \tfrac{\beta}{2\nu} \norm{\ell'(z^*)}^2 ,
  \end{align}
  where the first inequality uses the convexity of $\ell$ and the second one
  appies the strong convexity of $E$. The last line is obtained by closed form
  minimization over $z$. Rearranging the above yields
  \begin{align}
    \ell(z^*) \le \ell(\hat z) + \tfrac{1}{\beta} \big( E(\hat z;\theta) - E(z^*;\theta) \big) + \tfrac{\beta}{2\nu} \norm{\ell'(z^*)}^2 ,
  \end{align}
  Combining this with the lower bound finally provides
  \begin{align}
    \ell(\hat z) + \tfrac{1}{\beta} \big( E(\hat z;\theta) - E(z^*;\theta) \big) \le \ell(z^*)
    \le \ell(\hat z) + \tfrac{1}{\beta} \big( E(\hat z;\theta) - E(z^*;\theta) \big) + \tfrac{\beta}{2\nu} \norm{\ell'(z^*)}^2 ,
  \end{align}
  i.e.\ Eq.~\ref{eq:approximation}.
\end{proof}

\begin{proof}[Proof of Prop.~\ref{prop:bottom_up}]
  We start at the innermost level and obtain
  \begin{align}
    \min_\theta \ell(z_L^*) \qquad \text{s.t. } & z_k^* = \arg\min_{z_k} E_k(z_k; z_{k-1}^*; \theta) \qquad k=3,\dotsc,L \nonumber \\
    {} & z_2^* = \arg\min_{z_2} \left\{ \min_{z_1} \big\{ E_2(z_2,z_1;\theta) + \tfrac{1}{\beta_1} E_1(z_1;\theta) \big\}
         - \min_{z_1} \tfrac{1}{\beta_1} E_1(z_1;\theta) \right\} .
  \end{align}
  We define
  $\check E_1(\theta) := \min_{z_1} E_1(z_1;\theta)/\beta_1$. Hence the above
  reduces to
  \begin{align}
    \min_\theta \ell(z_L^*) \qquad \text{s.t. } & z_k^* = \arg\min_{z_k} E_k(z_k; z_{k-1}^*; \theta) \qquad k=3,\dotsc,L \nonumber \\
    {} & z_2^* = \arg\min_{z_2} \left\{ \min_{z_1} \big\{ E_2(z_2,z_1;\theta) + \tfrac{1}{\beta_1} E_1(z_1;\theta) \big\} - \check E_1(\theta) \right\}.
  \end{align}
  Applying the expansion on $z_3^*$ yields
  \begin{align}
    \min_\theta \ell(z_L^*) \qquad \text{s.t. } & z_k^* = \arg\min_{z_k} E_k(z_k; z_{k-1}^*; \theta) \qquad k=4,\dotsc,L \nonumber \\
    {} & z_3^* = \arg\min_{z_3} \bigg\{ \min_{z_2} \left\{ E_3(z_3,z_2; \theta)
         + \tfrac{1}{\beta_2} \min_{z_1} \big\{ E_2(z_2,z_1;\theta) + \tfrac{1}{\beta_1} E_1(z_1;\theta) \big\}
         - \check E_1(\theta) \right\} \nonumber \\
    {} &\hspace{6em}- \tfrac{1}{\beta_2} \min_{z_1,z_2} \big\{ E_2(z_2,z_1;\theta) + \tfrac{1}{\beta_1} E_1(z_1;\theta)
         - \check E_1(\theta)\big\} \bigg\} .
  \end{align}
  The $\check E_1(\theta)$ terms cancel, leaving us with
  \begin{align}
    \min_\theta \ell(z_L^*) \qquad \text{s.t. } & z_k^* = \arg\min_{z_k} E_k(z_k; z_{k-1}^*; \theta) \qquad k=4,\dotsc,L \nonumber \\
    {} & z_3^* = \arg\min_{z_3} \bigg\{ \min_{z_1,z_2} \left\{ E_3(z_3,z_2; \theta)
         + \tfrac{1}{\beta_2} E_2(z_2,z_1;\theta) + \tfrac{1}{\beta_1\beta_2} E_1(z_1;\theta) \right\} \nonumber \\
    {} &\hspace{6em}- \min_{z_1,z_2} \big\{ \tfrac{1}{\beta_2} E_2(z_2,z_1;\theta) + \tfrac{1}{\beta_1\beta_2} E_1(z_1;\theta) \big\} \bigg\}.
  \end{align}
  We define
  \begin{align}
    U_2(\theta) := \min_{z_1,z_2} \big\{ \tfrac{1}{\beta_2} E_2(z_2,z_1;\theta) + \tfrac{1}{\beta_1\beta_2} E_1(z_1;\theta) \big\}
  \end{align}
  and generally
  \begin{align}
    U_k(\theta) := \min_{z_1,\dotsc,z_k} \left\{ \sum_{l=1}^k \frac{1}{\prod_{j=l}^k \beta_j} E_k(z_k,z_{k-1};\theta) \right\} .
  \end{align}
  Thus, we obtain
  \begin{align}
    \min_\theta \ell(z_L^*) \qquad \text{s.t. } & z_k^* = \arg\min_{z_k} E_k(z_k; z_{k-1}^*; \theta) \qquad k=4,\dotsc,L \nonumber \\
    {} & z_3^* = \arg\min_{z_3} \left\{ \min_{z_1, z_2} \left\{ E_3(z_3,z_2; \theta)
         + \tfrac{1}{\beta_2} E_2(z_2,z_1;\theta) + \tfrac{1}{\beta_1\beta_2} E_1(z_1;\theta) \right\} - U_2(\theta) \right\} .
  \end{align}
  Expanding until $z_{L-1}^*$ yields
  \begin{align}
    \min_\theta \ell(z_L^*) \qquad \text{s.t. } z_L^* = \arg\min_{z_L} \left\{ \min_{z_1,\dotsc,z_{L-1}} \left\{
    \sum_{k=1}^{L} \frac{1}{\prod_{l=k}^{L-1}\beta_l} E_k(z_k,z_{k-1}; \theta) \right\} - U_{L-1}(\theta) \right\}
  \end{align}
  A final expansion step finally results in
  \begin{align}
  \min_\theta \left\{ \min_{z_1,\dotsc,z_L} \left\{ \ell(z_L) + \sum_{k=1}^{L} \frac{E_k(z_k,z_{k-1}; \theta)}{\prod_{l=k}^{L-1}\beta_l} \right\}
  - \min_{z_1,\dotsc,z_L} \left\{ \sum_{k=1}^{L} \frac{E_k(z_k,z_{k-1}; \theta)}{\prod_{l=k}^{L-1}\beta_l} \right\} \right\}
    = \min_\theta\, \cL_{GC}^{\bbeta}(\theta),
  \end{align}
  since the $U_{L-1}(\theta)$ terms cancel again.
\end{proof}

\begin{proof}[Proof of Prop.~\ref{prop:top_down}]
  We start at the outermost level $L$ and obtain
  \begin{align}
    \min_\theta \left( \min_{z_L} \big\{ \ell(z_L) + \tfrac{1}{\beta_L} E_L(z_L, z_{L-1}^*;\theta) \big\}
    - \min_{z_L} \tfrac{1}{\beta_L} E_L(z_L, z_{L-1}^*; \theta) \right) 
  \end{align}
  subject to $z_k^* = \arg\min_{z_k} E_k(z_k, z_{k-1}^*; \theta)$ for
  $k=1,\dotsc,L-1$. Using $z_L^* = f_L(z_{k-1}^*; \theta)$ and
  \begin{align}
    E_L(z_L^*, z_{L-1}^*;\theta) = E_L(f_L(z_{L-1}^*;\theta), z_{L-1}^*;\theta) =: \check E_L(z_{L-1}^*;\theta) 
  \end{align}
  the above objective reduces to
  \begin{align}
    \min_\theta &\left( \min_{z_L} \big\{ \ell(z_L) + \tfrac{1}{\beta_L} E_L(z_L, z_{L-1}^*;\theta) \big\}
                  - \tfrac{1}{\beta_L} \check E_L(z_{L-1}^*;\theta) \right) \nonumber \\
    = \min_\theta & \min_{z_L} \big\{ \ell(z_L) + \tfrac{1}{\beta_L} \big( E_L(z_L, z_{L-1}^*;\theta) - \check E_L(z_{L-1}^*;\theta) \big) \big\} 
  \end{align}
  subject to $z_k^* = \arg\min_{z_k} E_k(z_k, z_{k-1}^*; \theta)$ for
  $k=1,\dotsc,L-1$. We have a nested minimization with $L-1$ levels and a
  modified main objective. We apply the conversion a second time and obtain
  \begin{align}
    \min_\theta &\bigg( \min_{z_{L-1}}\left\{ \min_{z_L} \left\{ \ell(z_L) + \tfrac{1}{\beta_L} \big( E_L(z_L, z_{L-1};\theta)
                  - \check E_L(z_{L-1};\theta) \big) \right\} + \tfrac{1}{\beta_{L-1}} E_{L-1}(z_{L-1}, z_{L-2}^*; \theta) \right\} \nonumber \\
                &- \min_{z_{L-1}} \tfrac{1}{\beta_{L-1}} E_{L-1}(z_{L-1}, z_{L-2}^*; \theta) \bigg) \\
    = \min_\theta & \min_{z_{L-1}, z_L} \left\{ \ell(z_L)
                    + \tfrac{1}{\beta_L} \big( E_L(z_L, z_{L-1};\theta) - \check E_L(z_{L-1};\theta) \big)
                    + \tfrac{1}{\beta_{L-1}} \big( E_{L-1}(z_{L-1}, z_{L-2}^*; \theta) - \check E_{L-1}(z_{L-2}^*;\theta) \big) \right\} 
  \end{align}
  subject to the constraints on $z_k^*$ for $k=1,\dotsc,L-2$. By continuing this
  process we finally arrive at
  \begin{align}
    \min_\theta \min_{z_1,\dotsc,z_L} \left\{ \ell(z_L)
    + \sum\nolimits_{k=1}^L \frac{1}{\beta_k} \big( E_k(z_k, z_{k-1};\theta) - \check E_k(z_{k-1}; \theta) \big) \right\}
  \end{align}
  as claimed in the proposition.
\end{proof}

\begin{proof}[Prop.~\ref{prop:lin_top_down_step}]
  We introduce $g(z) := \min_y h(y,z)$ and $y^*(z) = \arg\min_y h(y,z)$, and observe that
  \begin{align}
    \frac{dg(z)}{dz} = \frac{\partial h(y^*(z), z)}{\partial z} + \frac{\partial h(y^*(z), z)}{\partial y}\cdot \frac{dy^*(z)}{dz}
    = \frac{\partial h(y^*(z), z)}{\partial z},
  \end{align}
  since $\partial_y h(y^*(z),z)=0$ by construction. This together with the
  definition of the linearized surrogate (Eq.~\ref{eq:lin_surrogate})
  yields the result.
\end{proof}

\begin{proof}[Proof of Prop.~\ref{prop:lin_top_down}]
  For brevity we omit the explicit dependence of $E$ on $\theta$ below.
  Applying the linearized contrastive surrogate on the outermost level of the
  deeply nested program in Eq.~\ref{eq:deepnested}
  yields
  \begin{align}
    \min_{z_L}\left\{ \ell(z_L^*) + (z_L-z_L^*)\tr \ell'(z_L^*) + \tfrac{1}{\beta_L} E_L(z_L,z_{L-1}^*) \right\}
    - \min_{z_L} \tfrac{1}{\beta_L} E_L(z_L,z_{L-1}^*) \qquad \text{s.t. } z_k^* = \arg\min_{z_k} E_k(z_k,z_{k-1})
    \label{eq:lin_first_step}
  \end{align}
  for $k=1,\dotsc,L-1$, where $\ell'(z_L) = \tfrac{d}{dz_L}\ell(z_L)$.
  Now we apply Prop.~\ref{prop:lin_top_down_step}
  with
  \begin{align}
    g_L^+(z_{L-1}) &:= \min_{z_L}\left\{ \ell(z_L^*) + (z_L-z_L^*)\tr \ell'(z_L^*) + \tfrac{1}{\beta_L} E_L(z_L,z_{L-1}) \right\}
    & g_L^-(z_{L-1}) &:= \min_{z_L} \tfrac{1}{\beta_L} E_L(z_L,z_{L-1})
  \end{align}
  and therefore
  \begin{align}
    \tfrac{\partial}{\partial z_{L-1}} g_L^+(z_{L-1}) = \tfrac{1}{\beta_L} \tfrac{\partial}{\partial z_{L-1}} E_L(\bar z_L,z_{L-1})
    & & \tfrac{\partial}{\partial z_{L-1}} g_L^-(z_{L-1}) = \tfrac{1}{\beta_L} \tfrac{\partial}{\partial z_{L-1}} E_L(z_L^*,z_{L-1}),
  \end{align}
  where $\bar z_L$ and $z_L^*$ are the respective minimizers,
  \begin{align}
    \bar z_L := \min_{z_L}\left\{ \ell(z_L^*) + (z_L-z_L^*)\tr \ell'(z_L^*) + \tfrac{1}{\beta_L} E_L(z_L,z_{L-1}) \right\}
    & & z_L^* := \arg\min_{z_L} \tfrac{1}{\beta_L} E_L(z_L,z_{L-1}).
  \end{align}
  Hence, the linearization of the outer loss in Eq.~\ref{eq:lin_first_step} at
  $z_{L-1}^*$ is given by
  \begin{align}
    z_{L-1} &\mapsto \min_{z_L}\left\{ \ell(z_L^*) + (z_L-z_L^*)\tr \ell'(z_L^*) + \tfrac{1}{\beta_L} E_L(z_L,z_{L-1}^*) \right\}
              - \min_{z_L} \tfrac{1}{\beta_L} E_L(z_L,z_{L-1}^*) \nonumber \\
    {} &+ (z_{L-1} - z_{L-1}^*)\tr \tfrac{1}{\beta_L} \tfrac{\partial}{\partial z_{L-1}} \big( E_L(\bar z_L, z_{L-1}^*) - E_L(z_L^*, z_{L-1}^*) \big) \\
    {} &= \ell(z_L^*) + (\bar z_L-z_L^*)\tr \ell'(z_L^*) + \tfrac{1}{\beta_L} E_L(\bar z_L,z_{L-1}^*) - \tfrac{1}{\beta_L} E_L(z_L^*,z_{L-1}^*) \nonumber \\
    {} &+ (z_{L-1} - z_{L-1}^*)\tr \tfrac{1}{\beta_L} \tfrac{\partial}{\partial z_{L-1}} \big( E_L(\bar z_L, z_{L-1}^*) - E_L(z_L^*, z_{L-1}^*) \big) \\
    {} &= \ell(z_L^*) + (\bar z_L-z_L^*)\tr \ell'(z_L^*) + \tfrac{1}{\beta_L} h_L(\bar z_L,z_{L-1}^*)
         + (z_{L-1} - z_{L-1}^*)\tr \tfrac{1}{\beta_L} \tfrac{\partial}{\partial z_{L-1}} h_L(\bar z_L, z_{L-1}^*)
  \end{align}
  using
  $h_k(z_k,z_{k-1}) := E_k(z_k,z_{k-1}) - \min_{z_k'} E_k(z_k',z_{k-1})$.
  Thus, after the second step of applying the linearized contrastive
  surrogates we arrive at
  \begin{align}
    & \ell(z_L^*) + (\bar z_L-z_L^*)\tr \ell'(z_L^*) + \tfrac{1}{\beta_L} h_L(\bar z_L,z_{L-1}^*) \nonumber \\
    {} &+ \min_{z_{L-1}} \left\{ (z_{L-1} - z_{L-1}^*)\tr \tfrac{1}{\beta_L} \tfrac{\partial}{\partial z_{L-1}} h_L(\bar z_L, z_{L-1}^*)
         + \tfrac{1}{\beta_{L-1}} E_{L-1}(z_{L-1},z_{L-2}^*) \right\} - \min_{z_{L-1}} \tfrac{1}{\beta_{L-1}} E_{L-1}(z_{L-1},z_{L-2}^*) \\
    {} &= \ell(z_L^*) + (\bar z_L-z_L^*)\tr \ell'(z_L^*) + \tfrac{1}{\beta_L} h_L(\bar z_L,z_{L-1}^*)
         + (\bar z_{L-1} - z_{L-1}^*)\tr \tfrac{1}{\beta_L} \tfrac{\partial}{\partial z_{L-1}} h_L(\bar z_L, z_{L-1}^*) + \tfrac{1}{\beta_{L-1}} h_{L-1}(\bar z_{L-1},z_{L-2}^*)
  \end{align}
  subject to $z_k^* = \arg\min_{z_k} E_k(z_k,z_{k-1})$ for
  $k=L-2,\dotsc,1$. Continuation of these steps until the innermost level
  yields the claim of Prop.~\ref{prop:lin_top_down}.
\end{proof}

\section{Bottom-Up Traversal for Linearized Surrogates}

Using the linearized surrogate (Eq.~\ref{eq:lin_surrogate}) is only possible
if $\ell(\cdot)$ is differentiable and each $E_k(z_k,z_{k-1})$ is
differentiable in $z_{k-1}$. Recall that
$z_k^* := \arg\min_{z_k} E_k(z_k,z_{k-1})$ (and
$z_1^* = \arg\min_{z_1} E_1(z_1;\theta)$). Proceeding analogously to
Prop.~\ref{prop:bottom_up} but using the linearized contrastive surrogates
yields a modified global contrastive objective, where the clamped and free
energies are suitable linearized versions of $\cE^{\bbeta}$ and
$\hat\cE^{\bbeta}$:
\begin{proposition}
  Let $\bbeta=(\beta_1,\dotsc,\beta_L)$ be positive spacing parameters. Applying
  the linearized contrastive surrogate (Eq.~\ref{eq:lin_surrogate})
  recursively on Eq.~\ref{eq:deepnested} from the innermost to the outermost
  level yields the following \emph{linearized global contrastive} objective,
  \begin{align}
    \label{eq:lin_GC}
    \bar\cL^{\bbeta}_{GC}(\theta) := \min_{z_1,\dotsc,z_L} \left\{ \ell(z_L^*) + (z_L-z_L^*)\tr\ell'(z_L^*) + \bar\cE^{\bbeta}(z;\theta) \right\}
    - \min_{z_1,\dotsc,z_L} \bar\cE^{\bbeta}(z;\theta),
  \end{align}
  where
  \begin{align}
    \bar\cE^{\bbeta}(z;\theta) := \sum_{k=2}^{L} \frac{E_k(z_k,z_{k-1}^*;\theta) + (z_{k-1}-z_{k-1}^*)\tr\partial_{z_{k-1}} E_k(z_k,z_{k-1}^*;\theta)}
    {\prod_{l=k}^{L}\beta_l} + \frac{E_1(z_1;\theta)}{\prod_{l=1}^{L}\beta_l}.
  \end{align}
\end{proposition}
In contrast to the global contrastive objective in Prop.~\ref{prop:bottom_up},
the resulting subproblem $\bar\cE^{\bbeta}(z;\theta)$ is generally not convex in
$z$ even when $\cE^{\bbeta}$ is. $z_k$ depends on both $z_{k-1}$ and
$z_{k+1}$ in $\bar\cE^{\bbeta}$, hence minimizing $\bar\cE^{\bbeta}$ (and its
loss-augmented variant) poses usually a difficult, non-convex optimization
problem in $z_1,\dotsc,z_L$. Consequently, the resulting linearized global
contrastive objective $\bar\cL^{\bbeta}_{GC}$ is of little further interest.

\section{Julia Source Code}

The plots in Fig.~1 in the main text were obtained by running the Julia code below.

\small
\begin{verbatim}
using Statistics, MLDatasets
using Flux, Flux.Optimise
using Flux: onehotbatch, onecold, crossentropy
using Base.Iterators: partition
using ChainRulesCore, Random

train_x, train_y = MNIST.traindata(); test_x, test_y  = MNIST.testdata()

train_X = Float32.(reshape(train_x, 28, 28, 1, :));
test_X = Float32.(reshape(test_x, 28, 28, 1, :));

train_Y = onehotbatch(train_y, 0:9); test_Y  = onehotbatch(test_y, 0:9)

########################################################################

# Configure settings
use_negative_init = false
use_FenBP = true

const beta = 0.01f0

########################################################################

my_relu(x) = relu(x)

function ChainRulesCore.rrule(::typeof(my_relu), x)
    z = my_relu(x)
    function pullback(dy)
        zz = my_relu(x - beta*dy)
        return NoTangent(), (z - zz) / beta
    end
    return z, pullback
end

# We use this definition to prevent Flux from optimizing away our custom AD rule.
function my_act(x; use_FenBP = use_FenBP)
    return use_FenBP ? map(my_relu, x) : relu.(x)
end

########################################################################

glorot_neg_uniform(rng::AbstractRNG, dims...) = (-abs.((rand(rng, Float32, dims...)
                                        .- 0.5f0) .* sqrt(24.0f0 / sum(Flux.nfan(dims...)))))
glorot_neg_uniform(dims...) = glorot_neg_uniform(Random.GLOBAL_RNG, dims...)

########################################################################

# Fix random seed for repeatability across runs
Random.seed!(42)

m = Chain(
    x -> reshape(x, :, size(x, 4)),
    Dense(28*28, 256, init = use_negative_init ? glorot_neg_uniform : Flux.glorot_uniform),
    my_act,
    Dense(256, 128), my_act,
    Dense(128, 64),  my_act,
    Dense(64, 10),   softmax
)

loss(x, y) = sum(crossentropy(m(x), y))
accuracy(x, y) = mean(onecold(m(x), 1:10) .== onecold(y, 1:10))

eta = 0.001f0; opt = ADAM(eta)

train = ([(train_X[:,:,:,i], train_Y[:,i]) for i in partition(1:50000, 50)])

epochs = 50
for epoch = 1:epochs
    iter = 1
    total_loss = 0.f0; train_accuracy = 0.f0

    for d in train
        ps = params(m)
        gs = gradient(ps) do
            l = loss(d...); total_loss += l; train_accuracy += accuracy(d...); l
        end
        update!(opt, ps, gs); iter += 1
    end

    total_loss /= iter; train_accuracy /= iter
    test_accuracy = accuracy(test_X, test_Y)

    println(epoch, " loss: ", total_loss,
            " (approx) train accuracy: ", 100*train_accuracy,
            " test accuracy: ", 100*test_accuracy,
            " (approx) train error: ", 100-100*train_accuracy,
            " test error: ", 100-100*test_accuracy)
end
\end{verbatim}

\end{document}